\documentclass[10pt,twocolumn,letterpaper]{article}

\usepackage[pagenumbers]{iccv} 

\usepackage[utf8]{inputenc} 
\usepackage[T1]{fontenc}    
\usepackage{url}            
\usepackage{booktabs}       
\usepackage{amsfonts}       
\usepackage{nicefrac}       
\usepackage{microtype}      
\usepackage{xcolor}         
\usepackage{picinpar}

\usepackage{booktabs}       
\usepackage{amsfonts}       
\usepackage{nicefrac}       
\usepackage{microtype}      
\usepackage{mathtools}
\usepackage{multirow}
\usepackage{bm}
\usepackage{epsfig}
\usepackage{graphicx}
\usepackage{caption}
\usepackage{subcaption}
\usepackage{amsthm}
\usepackage{amsmath}
\usepackage{amssymb}
\usepackage{enumitem}
\usepackage{makecell}
\usepackage{wrapfig}
\usepackage{indentfirst}
\usepackage{verbatim}
\usepackage{color}
\usepackage{setspace}
\usepackage{array}
\usepackage{booktabs}
\usepackage{stackengine}
\usepackage{algorithm}
\usepackage[algo2e,algoruled,boxed,lined]{algorithm2e}
\usepackage{algorithmicx}
\usepackage{graphicx}
\usepackage{xcolor}

\usepackage{caption}

\renewcommand{\vec}[1]{\boldsymbol{#1}}
\newcommand{\minisection}[1]{\vspace{.03in}\noindent{\textbf{#1}.}}

\definecolor{iccvblue}{rgb}{0.21,0.49,0.74}
\definecolor{citecolor}{HTML}{0071bc}
\usepackage[pagebackref,breaklinks,colorlinks,allcolors=iccvblue, linkcolor=red, citecolor=citecolor]{hyperref}

\title{Modular Customization of Diffusion Models via Blockwise-Parameterized Low-Rank Adaptation}

\def\spaces{~~~~~~}
\author{Mingkang Zhu\textsuperscript{1}\spaces{}Xi Chen\textsuperscript{3}\spaces{}Zhongdao Wang\textsuperscript{4}\spaces{} \\
Bei Yu\textsuperscript{1}\spaces{}Hengshuang Zhao\textsuperscript{3}\spaces{}Jiaya Jia\textsuperscript{2,5} \\ \\ \textsuperscript{1}~CUHK\spaces{}\textsuperscript{2}~HKUST\spaces{}\textsuperscript{3}~HKU\spaces{}\textsuperscript{4}~Huawei\spaces{}\textsuperscript{5}~SmartMore}

\begin{document}
\maketitle
\begin{abstract}
Recent diffusion model customization has shown impressive results in incorporating subject or style concepts with a handful of images.  However, the modular composition of multiple concepts into a customized model, aimed to efficiently merge decentralized-trained concepts without influencing their identities, remains unresolved. Modular customization is essential for applications like concept stylization and multi-concept customization using concepts trained by different users.  Existing post-training methods are only confined to a fixed set of concepts, and any different combinations require a new round of retraining. In contrast, instant merging methods often cause identity loss and interference of individual merged concepts and are usually limited to a small number of concepts. To address these issues, we propose BlockLoRA, an instant merging method designed to efficiently combine multiple concepts while accurately preserving individual concepts' identity. With a careful analysis of the underlying reason for interference, we develop the Randomized Output Erasure technique to minimize the interference of different customized models. Additionally, Blockwise LoRA Parameterization is proposed to reduce the identity loss during instant model merging. Extensive experiments validate the effectiveness of BlockLoRA, which can instantly merge 15 concepts of people, subjects, scenes, and styles with high fidelity.
\end{abstract}    
\section{Introduction}
\label{sec:intro}

\begin{figure}[tb!]
    \centering
    \includegraphics[width=0.48\textwidth]{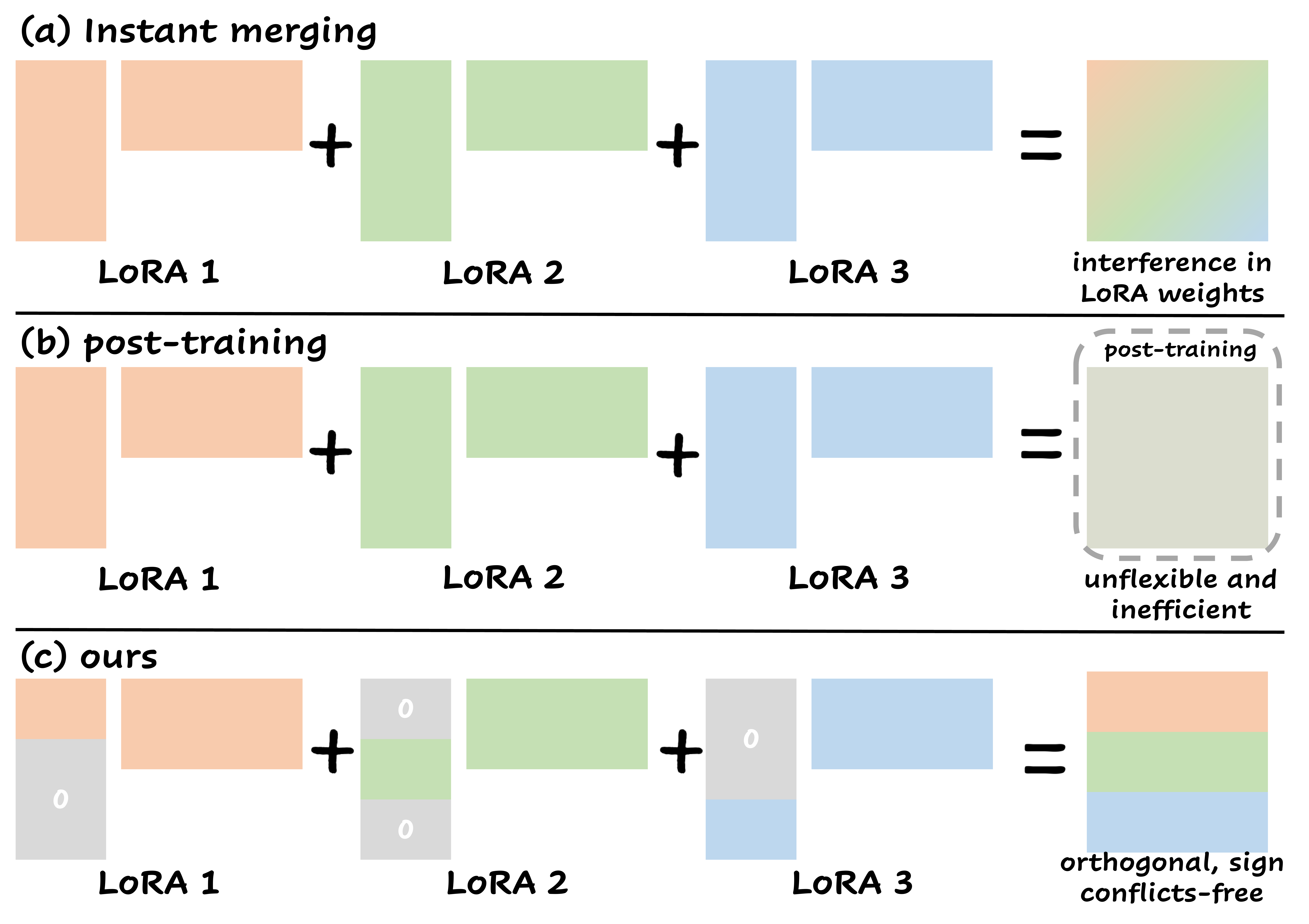}
    \caption{\textbf{Comparison of two main approaches} and ours for merging multiple concepts into a single pre-trained model. Our method supports instant merging while not hurting identities.}\label{model_comp}
\end{figure}

\noindent Text-to-image generative models have revealed impressive abilities in synthesizing high-quality images based on textual descriptions \cite{dalle2, dalle3,latent_diffusion,imagen,nichol2022glide,kang2023scaling,sauer2023stylegant,gafni2022makeascene}. Recently, a growing body of researches focuses on customized generation, which allows users to incorporate their own concepts into the pre-trained models and generate them in diverse scenarios \cite{Ruiz_2023_CVPR,gal2022textual, kumari2022customdiffusion, avrahami2023break, shah2023ziplora}. Although these customization methods can effectively insert a single  concept, merging multiple concepts into a single model is still challenging, especially for concepts modularly trained by different users. This capability is greatly sought-after and fundamental for applications like multi-concept customization and concept stylization. For example, users may wish to integrate multiple concepts trained by themselves or other users into a single model and explore interactions of any concept combinations. However, existing methods still fall short in handling this task \cite{po2023orthogonal, gu2023mixofshow,kumari2022customdiffusion, mcmahan2023communicationefficient}. Post-training methods \cite{gu2023mixofshow} are inflexible and time-consuming. Any different combinations of concepts would require new training from scratch. Also, the post-training time would increase with the number of concepts to be merged. On the other hand, instant merging methods \cite{mcmahan2023communicationefficient, po2023orthogonal} usually have issues with concept identity loss or interference. Although they can enable efficient merging, the number of merged concepts is usually small to avoid severe identity loss, as demonstrated in \Cref{model_comp}.

\begin{figure*}[h]
    \centering
    \includegraphics[width=1.0\textwidth]{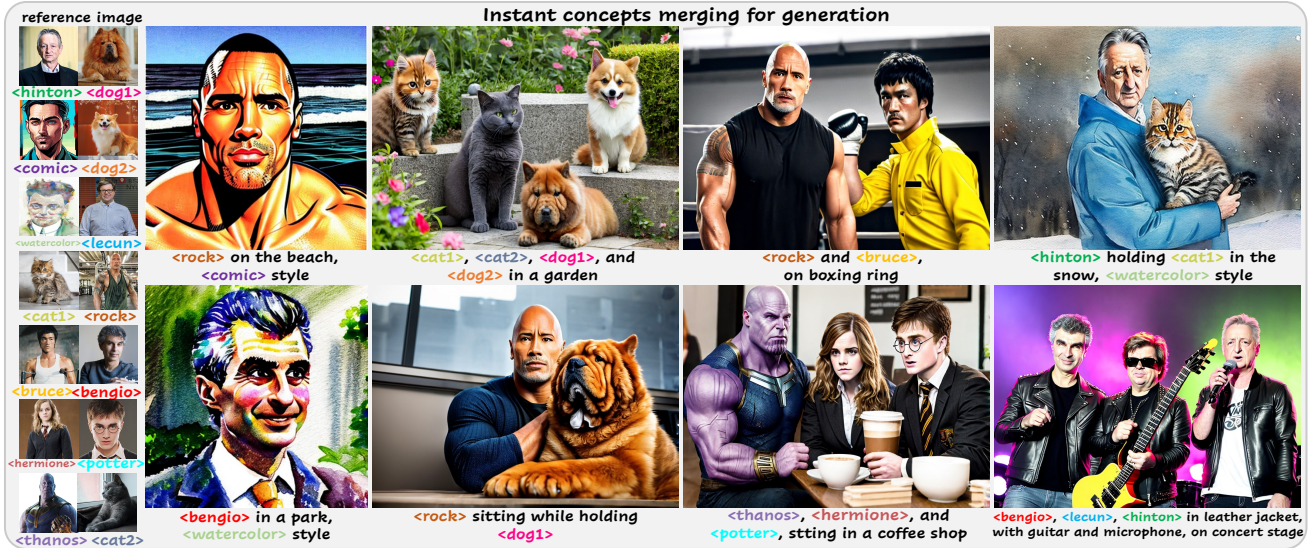}
    \caption{\textbf{Instant merging 15 concepts without hurting identities.} Given a set of concepts of characters, objects, scenes, or styles, our BlockLoRA can integrate any subsets of their identities instantly and accurately into pre-trained diffusion models without identity loss or interference. Thus meaningful interactions among merged concepts can be achieved smoothly.}
    \label{fig:tisser}
    
\end{figure*}

To design flexible, effective, and efficient methods to incorporate multiple concepts into diffusion models, the first and foremost thing is to understand the reasons behind concept identity interference and identity loss in merging customized models, which are often superficially perceived as conflicts in different models' weights \cite{gu2023mixofshow, po2023orthogonal}. A frequently encountered phenomenon of gradually diminishing prior preservation ability during diffusion model customization \cite{Ruiz_2023_CVPR}  gives us inspiration. 

Specifically, diffusion models being customized would gradually forget how to generate subjects of the same prior class as the user-specific concept. We  hypothesize that   the customization process is taking the easy route to modify the prior class's distribution instead of shaping an exclusive one for the user's concept.  In other words, the concept embeddings stay overly similar to the prior class token during the joint embedding-weight tuning, which unintentionally causes the distribution corresponding to the prior class token to drift. According to this hypothesis, merging customized models that share the same prior class could significantly disrupt the identities of the concepts, as each customized model learns its concept identity by altering the distribution of the same prior class. Such occurrences are not infrequent since users usually want to merge concepts of the same prior class, especially people, to synthesize group photos. The above analysis explains concept identity interference. Another challenge that hinders the effective merging of concepts is identity loss, which refers to the identities gradually growing dissimilar rather than mixing together. This is more related to the performance decay of merged customized models. Leveraging insights from model merging literature, we understand that the direction interference and shrinking magnitudes of the parameters to be merged are two major causes for performance decay \cite{task_arithmetic, yadav2023tiesmerging}. We identify that this issue also prevails in the case of merging customized models, which prior methods generally ignored \cite{gu2023mixofshow, po2023orthogonal}.

We propose BlockLoRA to address these issues. To alleviate the identity interference, we need to guide the customization process to create exclusive niches for each concept, instead of changing the prior class distribution. We draw inspiration from dropout and its variants \cite{dropout, dropblock, dropconnect}, which can mitigate overfitting by compelling the network to acquire redundant representations. We conjecture a comparable scenario for diffusion model customization. Therefore, we propose Randomized Output Erasure, which randomly erases residual LoRA outputs during the customization process. It introduces skewness into the output so that LoRA is forced to form a unique distribution for user-provided concepts instead of adjusting the prior class's distribution. To deal with the identity loss, we need to reduce the direction deviation and magnitude shrinking of the parameters to be merged. Specifically, the weight residuals added to the pre-trained models should:  \textcircled{\small{1}} exhibit proximity to orthogonality  \cite{task_arithmetic}, and \textcircled{\small{2}} demonstrate minimal sign conflicts \cite{yadav2023tiesmerging}. Because standard LoRA lacks such properties, we propose a novel Blockwise LoRA Parameterization so that each LoRA can only optimize a disjoint subset of parameters. We achieve this by introducing a disjoint binary mask for each LoRA. This formulation can naturally enable zero cosine similarity and no sign conflicts among different LoRAs. To reinforce the efforts to mitigate the identity interference problem, we propose to assign the binary masks as blocks of rows in the LoRA weight residual matrices, as this parameterization would also add skewness to the LoRA output pattern. 

With the aforementioned method, BlockLoRA is able to instantly insert 15 concepts into a single pre-trained diffusion model, spanning people, animals, subjects, scenes, and styles, without hurting each concept's identity. Being able to insert concepts into diffusion models like plug-and-play plugins, BlockLoRA seamlessly enables multi-concept customization and concept stylization, as shown in \Cref{fig:tisser}. Our contributions can be summarized as the following:
\begin{itemize}
  \item We conduct a thorough analysis of the reasons behind identity interference and identity loss, two major obstacles to the effective merging of customized models that are overlooked by previous approaches.
  
  \item Based on the careful analysis and correspondingly, we propose BlockLoRA, consisting of the Randomized Output Erasure and the Blockwise LoRA Parameterization. BlockLoRA minimizes the identity interference and identity loss for each concept and enables concept merging in a plug-and-play manner, allowing users to explore meaningful interactions among merged concepts.
  \item Comprehensive empirical study demonstrates the effectiveness of BlockLoRA over state-of-the-art customization methods, in terms of concept identity preservation ability, efficiency, and flexibility. 

\end{itemize}

\section{Related Work}
\label{Related Works}

\minisection{Text-to-image generative models} Extensive efforts have been made in text-to-image generative models in recent years \cite{latent_diffusion, imagen, dalle2, dalle3, nichol2022glide, sauer2023stylegant,kang2023scaling,gafni2022makeascene}, harnessing the advancements in diffusion models \cite{NEURIPS2020_4c5bcfec, song2021denoising,song2021scorebased,sohldickstein2015deep}, large-scale dataset \cite{laion}, and large vision-language models \cite{clip, blip}. Cutting-edge, large-scale models like Stable Diffusion \cite{latent_diffusion}, Imagen \cite{imagen}, and DALLE~3 \cite{dalle3} have empowered the generation of images with unparalleled fidelity and diversity following text prompts. Our method aims to integrate multiple concepts into these pre-trained models in a modular manner, without losing their unique characteristics. This allows users to investigate interactions among any combinations of these concepts flexibly.

\minisection{Diffusion model customization} Diffusion model customization aims to incorporate user-provided concepts into the pre-trained diffusion models and enable their contextualization in diverse scenarios. Each concept is usually represented using a few images. Textual Inversion \cite{gal2022textual} optimizes token embeddings to reconstruct the user-defined concepts. Dreambooth \cite{Ruiz_2023_CVPR}, on the other hand, fine-tunes the pre-trained models' weights using a reconstruction loss along with a prior preservation loss. Recent efforts try to use the low rank adaptation (LoRA) \cite{hu2022lora} for customizing diffusion models \cite{lora_diffusion}.  LoRA parameterization usually leads to better parameter efficiency and can still accurately capture the concept identity. Some works focus on fast or zero-shot customization of text-to-image diffusion models \cite{wei2023elite, shi2023instantbooth, jia2023taming, ruiz2023hyperdreambooth}. 
These methods usually require training a separate encoder on large-scale datasets to extract features of concepts.

Another line of research focuses on multi-concept customization, which aims to integrate multiple novel concepts into the pre-trained models \cite{kumari2022customdiffusion, gu2023mixofshow, po2023orthogonal, avrahami2023break, han2023svdiff, shah2023ziplora}. Custom Diffusion \cite{kumari2022customdiffusion} proposes to train multiple concepts simultaneously.
Break-a-Scene \cite{avrahami2023break}  and SVDiff \cite{han2023svdiff} also adopt the co-training approach and additionally introduce a masking strategy for better identity preservation. ZipLoRA \cite{shah2023ziplora} explores integrating one subject with one style at the same time. However, these methods can only work for at most 2-3 concepts and require the training data of all concepts concurrently,
which limits their efficiency and applicability.

Recent efforts try to extend multi-concept customization in a modular way, to improve its scalability and efficiency. Mix-of-show \cite{gu2023mixofshow} proposes the gradient fusion strategy, which post-trains the merged network to retain concept identities. The post-training approach is computationally expensive and is difficult to scale to a large number of concepts. It is also inflexible, as the post-trained weight only works for a fixed set of concepts. Any addition or removal of concepts would require a new post-training from scratch. Orthogonal Adaptation \cite{po2023orthogonal} enforces orthogonality for matrices $\vec{A}$s in LoRA weights $ \vec{\Delta W} = \vec{B}\vec{A}$ and develops a theory that ensures the disentanglement of concept identities in some hypothetical subspace spanned by the orthogonal complement of the rows of $\vec{A}$. However, the orthogonality for $\vec{A}$s does not result in the orthogonality of the LoRA weights $ \vec{\Delta W}$s added to pre-trained models. This can still lead to identity interference. Also, the feature dimensions in Stable Diffusion are usually not high enough to support strict orthogonality when combining multiple concepts. The study of Orthogonal Adaptation is only limited to merging 3 concepts at the same time \cite{po2023orthogonal}, which significantly hinders the applicability. While, our method allows instant merging for 15 LoRAs, and can better preserve the concept identities, especially when the number of merged concepts is large. \Cref{model_comp} summarizes the key difference between previous methods and ours.

\minisection{Model merging} Merging different fine-tuned models initialized from the same pre-trained weights has received increasing attention recently \cite{mcmahan2023communicationefficient, task_arithmetic, yadav2023tiesmerging, wortsman2022model, jin2023dataless}, as it provides an efficient way to expand the abilities of pre-trained models. Researchers have explored its capabilities in diverse areas, including multi-task learning \cite{wortsman2022model, task_arithmetic, yadav2023tiesmerging}, out-of-distribution generalization \cite{jin2023dataless}, and federated learning \cite{mcmahan2023communicationefficient}. \cite{wortsman2022model,mcmahan2023communicationefficient} propose to fuse fine-tuned models via simple weighted averaging. \cite{task_arithmetic} makes an extension by merging models with arithmetic operations, allowing for various merging effects. \cite{yadav2023tiesmerging} achieves effective fusion by filtering out parameters with sign conflicts. Driven by these advancements, we investigate the ideal method for merging customized diffusion models modularly.

\section{Method}
\label{Methods}
\noindent
In this section, we first provide a brief introduction of text-to-image diffusion models, customization, and problem setup in \Cref{Preliminary}.
Then we propose our BlockLoRA in \Cref{Randomized LoRA Output Erasure} and \Cref{block lora}.
In \Cref{Randomized LoRA Output Erasure}, we thoroughly examine the underlying reason behind identity interference and propose Randomized Output Erasure to address the issue.
Finally, we analyze the cause of identity loss from a model merging perspective and propose a novel LoRA architecture as a remedy in \Cref{block lora}.

\subsection{Preliminary}
\label{Preliminary}

\minisection{Diffusion models}
Diffusion models \cite{NEURIPS2020_4c5bcfec, song2021denoising, song2021scorebased} are a type of generative models that approximate data distributions by progressively removing noise from samples of a Gaussian distribution. They are usually designed to operate in the latent space for efficiency \cite{latent_diffusion, dalle3}. With a pre-trained autoencoder mapping images $\vec{x}$ to compressed latent space, diffusion models are trained to denoise the encoded latents $\vec{z}$. Diffusion models can be conditioned on a conditioning vector $\vec{c} = \gamma_{\vec{\theta}}(P)$, where $\gamma_{\vec{\theta}}$ is a pre-trained text encoder and $P$ is the text prompt. Then the diffusion loss is:
\begin{equation*}
\begin{aligned}
L = \mathbb{E}_{\vec{z}, P, t, \vec{\epsilon} \sim \mathcal{N}(0,\,1)} \left[ ||\vec{\epsilon}-\vec{\epsilon_\theta}(\vec{z}_t, t, \gamma_{\vec{\theta}}(P))||_2^2 \right],
\end{aligned}
\end{equation*}
where $\vec{\epsilon}$ is the added Gaussian noise, $t$ is timestep, $\vec{z}_t$ is the latent representation at $t$, and $\vec{\epsilon_\theta}$ is the denoising U-Net \cite{ronneberger2015unet}.

\minisection{Low-rank adaptation}
Low-rank adaptation (LoRA) \cite{hu2022lora} is a parameter-efficient fine-tuning method 
built upon a hypothesis that the weight changes during model fine-tuning have a low ``intrinsic rank''. LoRA freezes the pre-trained model weights and constrains the updates in a low rank decomposition of the original weight matrix. Specifically, for a weight matrix \( \vec{W}_0 \in \mathbb{R}^{m \times n} \) of the pre-trained model, LoRA obtains the updated weight $\vec{W}$ by \(\vec{W} = \vec{W}_0 + \vec{\Delta W}\). The weight residual \(\vec{\Delta W} = 
\vec{B}\vec{A} \), where \( \vec{B} \in \mathbb{R}^{m \times r} \) and \( \vec{A} \in \mathbb{R}^{r \times n} \) are trainable low-rank matrices, with the rank \( r \ll \min(m, n) \) to enforce the low rank constraint. With $\vec{x}$ as the input, the output $\vec{h} = \vec{W}_0 \vec{x}$ is modified as:
\begin{equation}
\label{lora_output}
\begin{aligned}
\vec{h} = \vec{W}_0 \vec{x} + \vec{\Delta W} \vec{x} = \vec{W}_0 \vec{x} + \vec{B}\vec{A} \vec{x}.
\end{aligned}
\end{equation}
LoRA has been widely adopted for customizing diffusion models \cite{lora_diffusion} with novel concepts. When multiple LoRAs contain different concepts, simple weighted averaging is commonly used to merge multiple LoRAs into one \cite{mcmahan2023communicationefficient, wortsman2022model}. Formally, when combining $k$ distinct LoRAs, the updated weight $\vec{W}$ is obtained by:
\begin{equation}
\label{fed_avg}
\begin{aligned}
\vec{W} = \vec{W}_0 + \sum_{i=1}^k \alpha_i \vec{\Delta W}_i, \quad \text{s.t.} \quad \sum_{i=1}^k \alpha_i = 1,
\end{aligned}
\end{equation}
where $\vec{\Delta W}_i$ is the $i$-th LoRA weight and $\alpha_i$ is the corresponding nonnegative scalar coefficient.

\minisection{Problem setup}
Given a set \(\mathbb{S}\)   of LoRAs, each representing a distinct concept, we aim to enable instant merging of any subset $\mathbb{T}=\{\vec{\Delta W}_i, i = 1, 2, \ldots, n\} \subseteq \mathbb{S}$ into the pre-trained diffusion model $\vec{W}_0$. The identity of each concept in $\mathbb{T}$ should be accurately preserved after merging. In this way, meaningful interactions among decentralized-trained concepts by different users, like multi-concept customization or concept stylization, can be achieved seamlessly.

\subsection{Randomized Output Erasure}
\label{Randomized LoRA Output Erasure}
\noindent
Although simple weighted averaging can efficiently fuse any combinations of LoRAs, this method usually leads to severe identity entanglement of the merged concepts \cite{wu2024mixture, gu2023mixofshow, po2023orthogonal}. Prior studies tentatively attribute this phenomenon to the conflicts in different LoRA weights, without further analysis of its underlying causes \cite{gu2023mixofshow,po2023orthogonal}. Inspired by the commonly observed phenomenon of diminishing prior preservation ability during diffusion model customization \cite{Ruiz_2023_CVPR}, we hypothesize the underlying cause for the identity interference of merged concepts is that different LoRAs modify the same prior class distribution instead of creating their own niches. Specifically, the diminishing prior preservation ability refers to the fact that the fine-tuned diffusion model loses the ability of generating subjects of the same class as the user-provided concept. We hypothesize that this is due to the fine-tuning process taking the ``shortcut'' to shape the prior class distribution instead of crafting one specific to the user-provided concept. In other words, the concept embeddings remain too similar to the prior class token during the joint embedding-weight tuning, which unintentionally leads to the drift of distribution corresponding to the prior class token. This problem would be more pronounced as users usually want to merge multiple concepts with the same prior class, like people or pets, to generate group photos. 

\begin{figure}[t]
  \centering
  \includegraphics[width=0.48\textwidth]{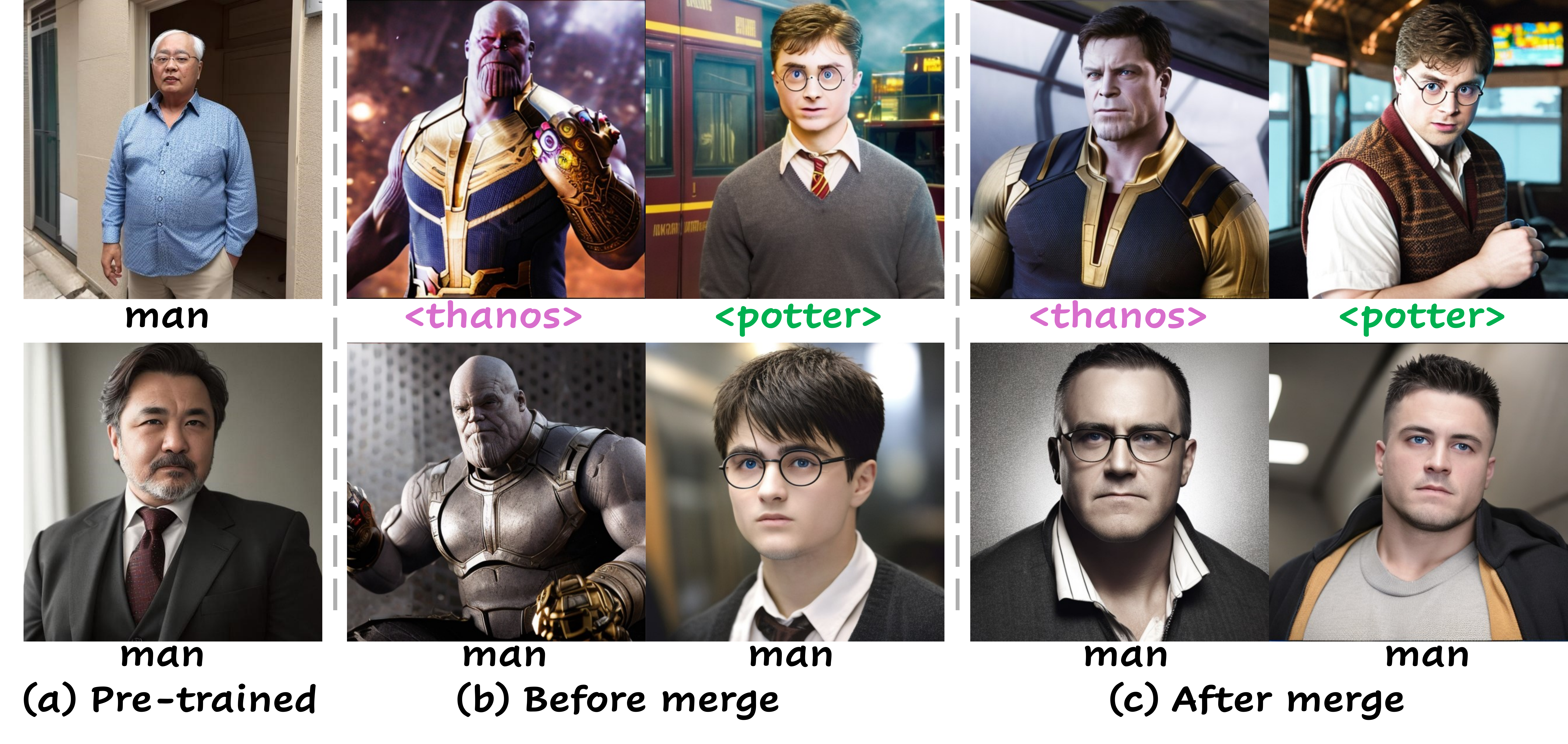}
  \caption{\textbf{Visualization of prior class distribution drift:} (a) Pre-trained model's prior class generations; (b) Customized models' concept generations and prior class generations; (c) Merged model's concept generations and prior class generations. The prior class generations are significantly drifted towards concepts after fine-tuning. Both the prior class generations and concept identities are entangled after merging.}
  \label{fig:motivation_1}
\end{figure}

To better demonstrate our hypothesis, we conduct an experiment in \Cref{fig:motivation_1} where we fine-tune two models customizing Thanos and Harry Potter, and merge them using simple weighted averaging.
In \Cref{fig:motivation_1}(a), we show the prior class generations of the pre-trained model, which is ``man'' in this case.
We then show the concept generations and the prior class generations for individual fine-tuned models and the merged model in \Cref{fig:motivation_1}(b) and \Cref{fig:motivation_1}(c) correspondingly.
From \Cref{fig:motivation_1}, we can see that the individual fine-tuned models' generations of ``man'' become extremely similar to the concepts been customized. After merging, the generations of Harry Potter and Thanos display characteristics of both characters and so do the generations of class prior ``man''. These results validate our hypothesis that the fine-tuning process takes the ``shortcut'' to modify the prior class distribution.

To alleviate this issue, we draw inspiration from the dropout technique and its variants \cite{dropout, dropblock, dropconnect}, which can prevent overfitting by forcing the network to learn redundant representations. We hypothesize a similar scenario for diffusion model customization, where we can force learning the concept's own distribution instead of relying on drifting the existing prior class distribution. Therefore, we propose to randomly erase the residual part of the LoRA output. In this way, we add distortion to the LoRA residual's output pattern, making it difficult to take advantage of the prior class distribution in the pre-trained model.
Specifically, the LoRA output $\vec{h} = \vec{W}_0 \vec{x} + \vec{B}\vec{A} \vec{x}$ in \Cref{lora_output} is modified as:
\begin{equation}\label{eraser}
\begin{aligned}
 \vec{h} = \vec{W}_0 \vec{x} +  \vec{r}(\lambda) \odot \vec{\Delta W} \vec{x} = \vec{W}_0 \vec{x} + \vec{r}(\lambda) \odot \vec{BA x},
\end{aligned}
\end{equation}
where $\odot$ denotes the element-wise product and $\vec{r}(\lambda)$ is a binary mask of size $m$, with each element drawn independently from $m_i \sim Bernoulli(1-\lambda)$. With the randomized output erasure, each customized concept can reside in its own niche, which provides a beneficial starting point for subsequent model merging.

\subsection{Blockwise LoRA Parameterization}
\label{block lora}

\begin{figure}[tb!]
  \centering
  \includegraphics[width=0.96\linewidth]{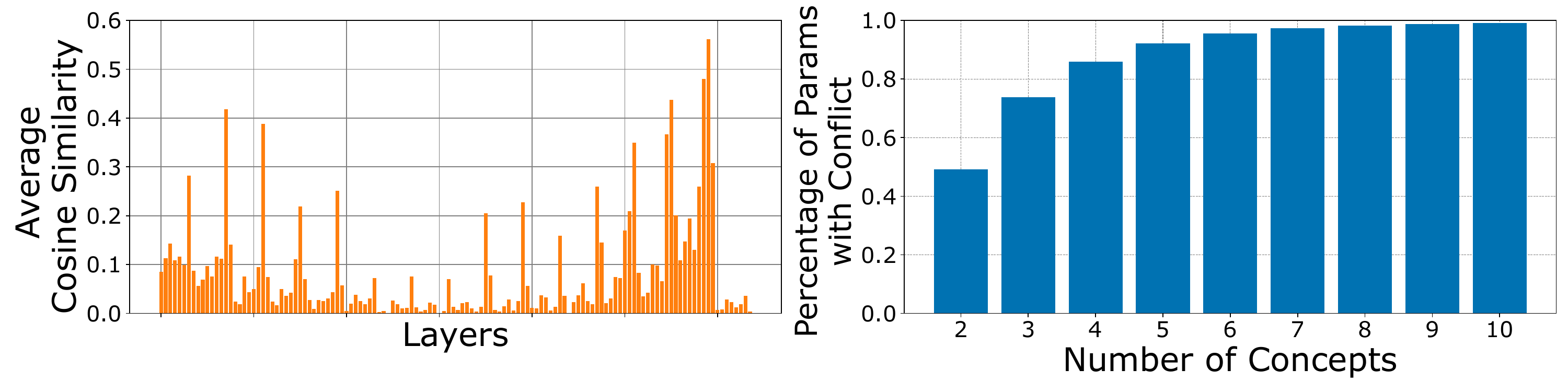}
  \caption{\textbf{Visualization of direction interference and sign conflicts.} Left: Layer-wise average cosine similarity for every combination of models in 10 customized models; Right: Fraction of parameters with sign conflicts when the number of customized models increases from 2 to 10.}
  \label{fig:motivation_2}
\end{figure}

\noindent
As we aim for the instant merging of LoRAs, we stick to the weighted averaging approach and address its limitations.  Leveraging insights from model merging literature, we know that weighted averaging would cause direction drift and diminishing magnitudes of the parameters, thereby leading to worse performance of the merged model \cite{task_arithmetic, yadav2023tiesmerging}. Specifically, for different fine-tuned models to be merged more effectively, the weight residuals added to the pre-trained weight for different models should: \textcircled{\small{1}} be close to orthogonal \cite{task_arithmetic}, and \textcircled{\small{2}} have few sign conflicts \cite{yadav2023tiesmerging}. We start by investigating whether LoRAs fine-tuned on different concepts exhibit these properties. Specifically, we fine-tune 10 LoRAs on different concepts and visualize the layer-wise average cosine similarity for every LoRA combination, and the fraction of parameters with sign conflicts when merging 2 to 10 models.
As shown in \Cref{fig:motivation_2}, we can see that many LoRA layers do not have near-zero cosine similarity. Some LoRA layers even have cosine similarity as high as 0.3.
Moreover, from \Cref{fig:motivation_2}, we can see that about half of the parameters have sign conflicts for only two models. Notably, nearly all parameters have sign conflicts when merging 10 models. This can be detrimental to accurately preserving the concept identity stored in each LoRA. 

Since standard LoRA does not have the above-mentioned two nice properties, we propose to modify its structure to address this issue.  We propose to apply a binary mask to each LoRA's $\vec{\Delta W} = \vec{B}\vec{A}$, so that each LoRA can only optimize a disjoint subset of parameters. In this way, the cosine similarity between any two LoRAs would naturally be zero and there will also be no sign conflicts because each LoRA allocates its parameters in different entries in the weight matrix $\vec{\Delta W}$. 

There can be numerous ways of selecting disjoint binary masks for different LoRAs. For simplicity, the binary masks can be chosen to assign 1 only to the entries of a subset of columns or rows in weight matrix $\vec{\Delta W}$.
Since we discussed in \Cref{Randomized LoRA Output Erasure} that adding distortion to the LoRA residual's output pattern would be beneficial for minimizing the interference of different customized concepts, we propose to make each LoRA to be a block of rows in the weight matrix $\vec{\Delta W}$, to strengthen the distortion. Formally, for all $q$, 
\begin{gather} 
 \vec{\Delta W}_i =  (\vec{M}_i \odot \vec{B}_i) \vec{A}_i, \label{iLoRA} \\ 
  \vec{M}_i^{pq} = \begin{cases}
    1,& \text{if } p \in \mathbb{R}_i;\\
    0,              & \text{otherwise,}
\end{cases}  \nonumber \\
\mathbb{R}_i \cap \mathbb{R}_j = \emptyset,  \textrm{ for all }  i \neq j, \nonumber
\end{gather}
where $\vec{\Delta W}_i$ is the residual parameter for the $i$-th LoRA, $\vec{M}_i$ is a binary mask and  $\vec{M}_i^{pq}$ is the entry in the $p$-th row and $q$-th column of the matrix $\vec{M}_i$, $\mathbb{R}_i$ is the subset of rows assigned for the $i$-th LoRA. $\mathbb{R}$s are constrained to be pairwise disjoint to ensure zero cosine similarity and zero sign conflicts for any LoRA combinations.

\begin{table*}[tb!]
    \footnotesize
\renewcommand{\arraystretch}{1.0}
\centering \caption{\textbf{Quantitative comparison} between our BlockLoRA and baselines, including merging time, text alignment, and image alignment. Text alignment and image alignment are calculated for each method after merging all 15 concepts.} \label{tab1}
\setlength{\tabcolsep}{11pt}
\begin{tabular}{l c c c c c}
\Xhline{1pt}
{\small Merge Type} & {\small  Method}  & {\small Merge Time} & {\small Image Align$\uparrow$} & {\small Text Align$\uparrow$}  & {\small Avg$\uparrow$}\\
\hline
{\small Post-Training}  & {\small Mix-of-show \cite{gu2023mixofshow} gradient fusion}   & {\small  {\color{red}$\sim 85$ m} }  & {\small  $0.780$ } &  {\small $0.724$ } &  {\small $0.752$ }\\
\hline
\multirow{4}{*}{ \small  Instant} &  {\small DB-LoRA \cite{hu2022lora, Ruiz_2023_CVPR}}  & {\small  $< 1$ s}  & {\small $0.703$} &  {\small $0.760$} &  {\small $0.732$}\\
 & {\small Mix-of-show \cite{gu2023mixofshow} weighted averaging}  & {\small  $< 1$ s}  & {\small $0.719$ } & {\small  $0.749$ } & {\small  $0.734$ } \\
 & {\small   Orthogonal Adaptation  \cite{po2023orthogonal} }    & {\small  $< 1$ s}  &  {\small $0.749$} & {\small $0.736$}  & {\small $0.743$}\\
 & {\small \textbf{Ours} }  & {\small $< 1$ s}   &  {\small  $0.775$ } & {\small  $0.743$} & {\small  $0.759$}\\

\Xhline{1pt}
\end{tabular}
\end{table*}

Thus, combing the Randomized Output Erasure in \Cref{eraser} and the Blockwise LoRA Parameterization in \Cref{iLoRA},
the $i$-th LoRA parameters during training are $\vec{\Delta W}_i =  \vec{r}(\lambda_i) \odot  (\vec{M}_i \odot \vec{B}_i) \vec{A}_i$.
The binary mask $\vec{r}(\lambda)$ is only active during training as a regularization and is removed once the training is done. $\vec{M}_i$ is kept as we need it to ensure zero sign conflicts and orthogonality among LoRAs. Consequently, during the customization process of the $i$-th LoRA, the customized model weight is defined as the following:

\begin{equation*} 
\begin{aligned}
\vec{W} = \vec{W}_0 + \vec{r}(\lambda_i) \odot   (\vec{M}_i \odot \vec{B}_i) \vec{A_i}. 
\end{aligned}
\end{equation*}

Then, for the given set \(\mathbb{T}=\{ \vec{\Delta W}_i, i = 1, 2, \ldots, n\}\)  of $n$ trained LoRAs, we use the following method to instantly merge LoRAs in $\mathbb{T}$ into the pre-trained diffusion model $\vec{W}_0$: 

\begin{equation*} 
\begin{aligned}
\vec{W} = \vec{W}_0 + \sum_{i=1}^n \alpha_i [(\vec{M}_i \odot \vec{B}_i) \vec{A}_i], \quad \text{s.t.} \quad \sum_{i=1}^n \alpha_i = 1.
\end{aligned}
\end{equation*}

With disjoint $\vec{M}_i$ in the above equation, the cosine similarity between different LoRAs is automatically zero, and there will also be no sign conflicts.

\section{Experiment}
\label{Experiment}
\noindent
In this section, we demonstrate the experiment setup, including the dataset we used, the evaluation metrics, and state-of-the-art baseline methods. Then we compare our BlockLoRA with state-of-the-art baselines quantitatively and qualitatively. Finally, we present the ablation study.

\subsection{Experimental Setup}
\label{Experimental Setup}

\minisection{Dataset}
We collect a dataset consisting of 15 concepts, spanning people, subjects, styles, and scenes to ensure diversity. Each concept is tested using 20 prompts. Existing baseline methods generally use similar amounts of test concepts. Mix-of-show \cite{gu2023mixofshow} adopts 14 real-world concepts while Orthogonal Adaptation \cite{po2023orthogonal} uses a total of 12 concepts.

\minisection{Baselines} We compare our BlockLoRA with 4 state-of-the-art baseline methods: Dreambooth LoRA (DB-LoRA) \cite{hu2022lora, Ruiz_2023_CVPR} with weighted averaging, Mix-of-show \cite{gu2023mixofshow} with  weighted averaging, Mix-of-show  \cite{gu2023mixofshow} with gradient fusion, and Orthogonal Adaptation \cite{po2023orthogonal}. For each method, we first fine-tune all concepts individually. Then we conduct the evaluation of the model containing all 15 merged concepts.

\minisection{Evaluation metrics} Following prior practices \cite{Ruiz_2023_CVPR,gal2022textual,gu2023mixofshow,po2023orthogonal}, we evaluate all methods using two metrics: text alignment and image alignment. Text alignment measures the similarity between the text prompts and the generations, while image alignment measures the similarity between the generations and reference images. Both image alignment and text alignment are calculated using CLIP scores \cite{clip}. 

\subsection{Comparison}

\begin{figure}[t]
    \centering  
    \includegraphics[width=0.9\linewidth]{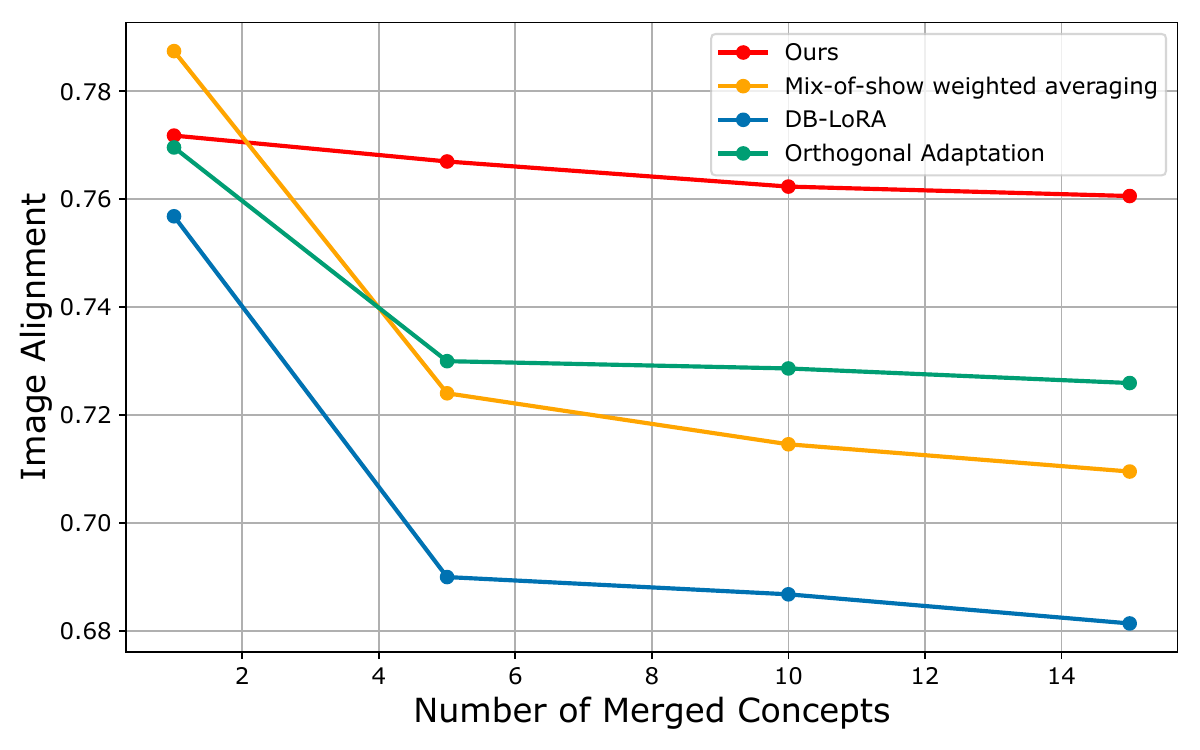}
    \caption{
        \textbf{Comparison of identity preservation ability} for instant merging methods when the number of merged concepts increases from 1 to 15.
        Our method only has marginal identity loss while others suffer greatly.
    }
    \label{fig:num_concept}
\end{figure}

\begin{figure*}[tb!]
    \centering
    \includegraphics[width=\textwidth]{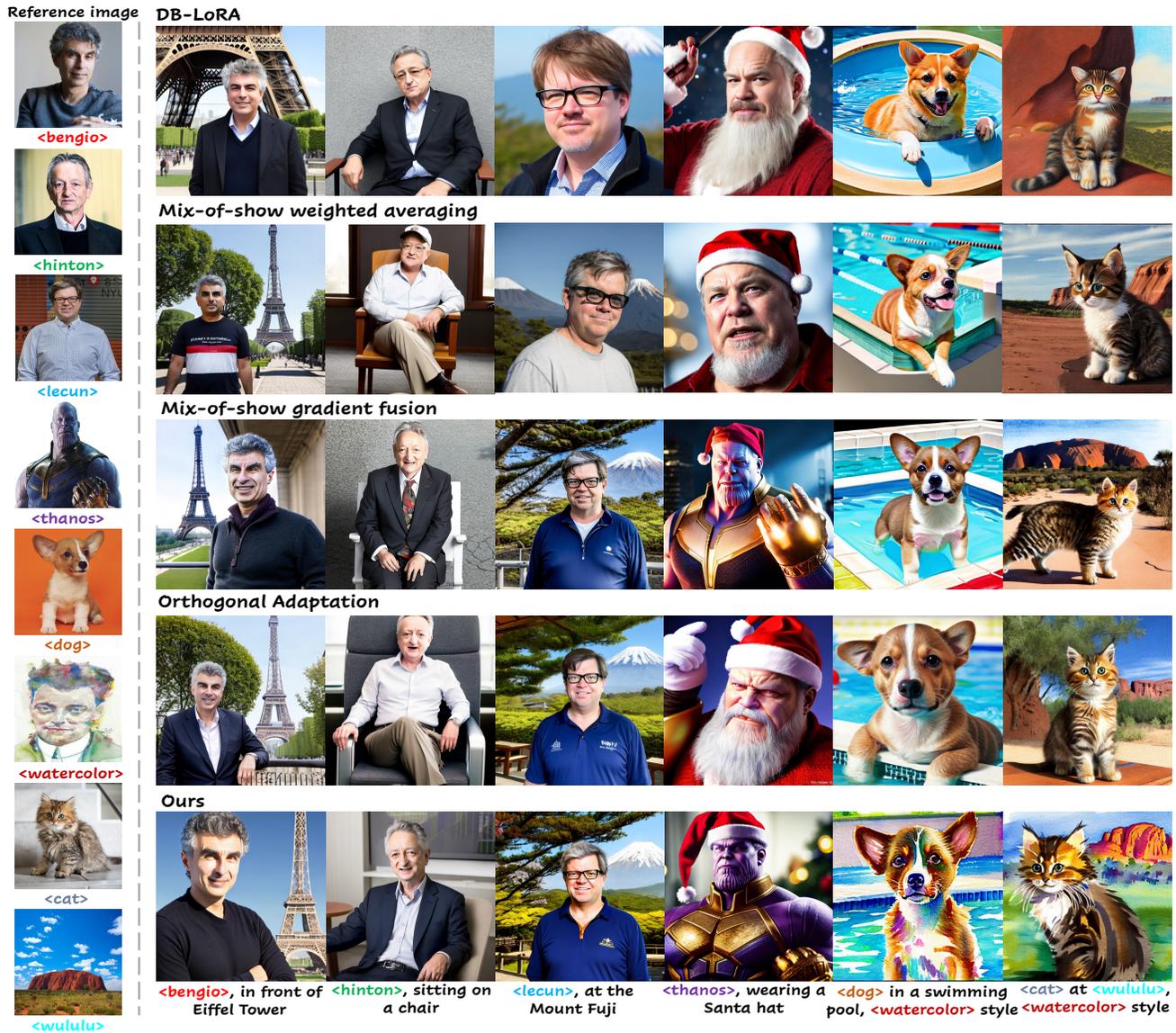}
    \caption{\textbf{Qualitative comparisons with baseline methods}. The visualizations are generated with all 15 customized concepts merged into the pre-trained model. Our BlockLoRA can accurately preserve the concept identities of instantly merged concepts, while others cannot. Our BlockLoRA can also support stylization with user-provided subject and style concepts, which is not applicable to other methods.}
    \label{fig:comp_single}
    \vspace{-0.3cm}
\end{figure*}

\begin{table}[tb!]
    \footnotesize
    \renewcommand{\arraystretch}{1.0}
    \centering
    \caption{\textbf{Users' preference on generations} by different methods.}
    \label{user-study}

    {
    \begin{tabular}{l c}
        \Xhline{1pt}
        Method & User Preference\\
        \hline
        Orthogonal Adaptation \cite{po2023orthogonal} &17\%\\

        Mix-of-show \cite{gu2023mixofshow} weighted averaging &2\%\\

        Mix-of-show \cite{gu2023mixofshow}  gradient fusion &39\%\\

        {\bf Ours} &42\%\\
        \Xhline{1pt}
    \end{tabular}
    }
\end{table}

\minisection{Quantitative results}
\Cref{tab1} gives quantitative comparisons of image alignment and text alignment between BlockLoRA and baselines. 
From \Cref{tab1}, we can see that our BlockLoRA achieves much higher image alignment than baseline instant merging methods. Our text alignment also remains comparable to other instant merging methods. This indicates that our method can preserve the concept identity much better than all baseline instant merging methods while maintaining a similar level of prompt-following ability. When comparing with post-training methods, we can see that our instant BlockLoRA achieves comparable image alignment and higher text alignment than Mix-of-show gradient fusion, which requires 85 minutes post-training.
In \Cref{fig:num_concept}, we visualize the instant merging methods’ identity preservation ability when the number of merged concepts increases from 1 to 15. We can see that our method has the least identity loss among all instant merging methods. Other instant merging methods generally have drastic identity loss as the number of merged concepts increases. This further validates our method's capability of merging a large number of concepts.
We also conduct a human evaluation in \Cref{user-study}, which shows that users prefer our method over others.

\begin{figure*}[t!]
  \centering
  \includegraphics[width=\textwidth]{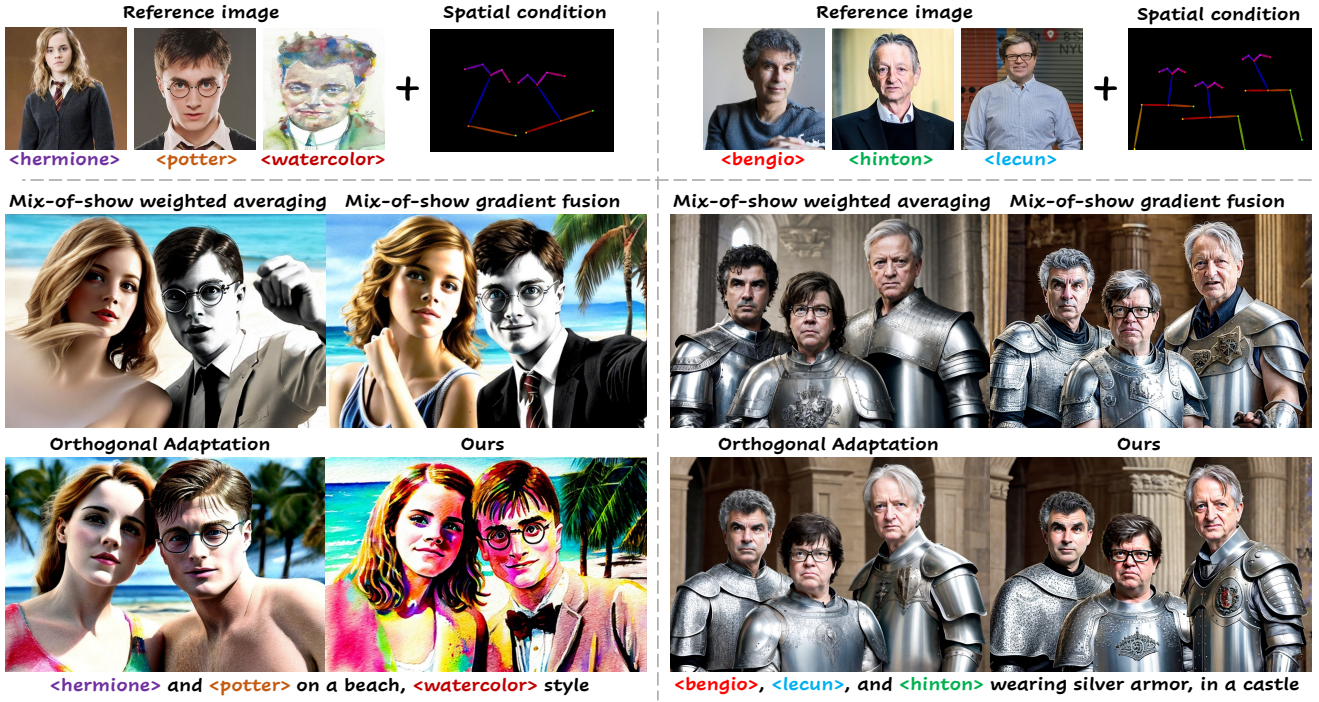}
  \caption{\textbf{Qualitative comparisons with baseline methods using spatial conditions}. On the left, our BlockLoRA can achieve multi-concept stylization while all baseline methods struggle and have severe identity loss. On the right, our BlockLoRA suffers the least identity loss while not requiring computationally intensive post-training. Especially, Lecun's haircut is changed for instant merging baselines, while our method maintains its identity.}
  \label{fig:comp_multi}
\vspace{-0.2cm}  
\end{figure*}

\minisection{Qualitative results}
We present the qualitative comparisons in \Cref{fig:comp_single},
where we can see that BlockLoRA can accurately preserve the concept identities for instantly merged concepts,
while other instant merging methods have different levels of identity loss.
Especially, we can see that all baseline methods except Mix-of-show gradient fusion fail to maintain the identity of Thanos, while BlockLoRA successfully handles this case.
Also, we notice from the last two columns of \Cref{fig:comp_single} that only our BlockLoRA is capable of performing concept stylization with customized subjects and styles. Our method can harmoniously integrate the features from both the subject concepts and style concepts, which further demonstrates its applicability.

\minisection{Qualitative results using spatial conditions}  We then present the comparisons with baselines for customization using spatial conditions in \Cref{fig:comp_multi}.
The results of DB-LoRA are omitted due to its poor performance.
From the left part of \Cref{fig:comp_multi}, we can see that BlockLoRA is also able to perform concept stylization for multiple subject concepts, while the baseline methods struggle and have severe identity loss. On the right, we can see that our method leads to better image alignment while other methods generally have identity loss. Mix-of-show gradient fusion can preserve the concept identities, but it is computationally unfriendly, especially for merging a large number of concepts.

\subsection{Ablation Study}
\noindent
In this section, we conduct an ablation study on the effect of each component of our BlockLoRA. We first examine the effect of the Randomized Output Erasure.
We can see from row 1 of \Cref{ablation} that the image alignment drops significantly without the randomized output erasure.
This further validates our hypothesis in \Cref{Randomized LoRA Output Erasure} that the customization process tends to take the shortcut to shape the prior class distribution. Therefore, the Randomized Output Erasure is crucial for better merging of customized models. 
 
\begin{table}[tb!]
    \renewcommand{\arraystretch}{1.0}
    \footnotesize
    \centering
    \caption{\textbf{Ablation study.} Quantitative analysis on using: (1) Randomized Output Erasure; (2) Blockwise LoRA Parameterization.}
    \label{ablation}
    {
        \begin{tabular}{l c c c}
            \Xhline{1pt}
            Method & Image Align$\uparrow$ &Text Align$\uparrow$ &Avg$\uparrow$\\
            \hline
            w/o  Output Erasure &0.721&0.713 &0.717\\

            w/o Blockwise  Para. &0.712&0.733&0.723\\

            {\bf Ours} &0.757&0.709&0.733\\
            \Xhline{1pt}
        \end{tabular}
    }
\end{table}

We then ablate the effect of our Blockwise LoRA Parameterization.
We can see from row 2 of \Cref{ablation} that the image alignment even drops more drastically.
This is because we are merging significantly more concepts than previous instant merging attempts and the identity loss caused by sign conflicts and high cosine similarity would be severe, as we discussed in \Cref{block lora}.
Therefore, the Blockwise LoRA Parameterization is necessary for minimizing identity loss.

\section{Conclusion}
\noindent
This paper explores how to merge concepts into pre-trained diffusion models instantly and effectively. We start by thoroughly examining the underlying causes for identity loss and interference of the merged concepts. We identify the cause of identity interference to be different LoRAs modifying the same prior class distribution. We also locate the reason for identity loss to be sign conflicts and direction deviations of LoRA parameters. According to our analysis, we propose BlockLoRA, consisting of the Randomized Output Erasure to prevent harmful shaping of the prior class distribution and the Blockwise LoRA Parameterization to enforce zero sign conflicts and orthogonality among all LoRAs. BlockLoRA enables the instant merging of 15 concepts without hurting their identities. Extensive experiment demonstrates the effectiveness of BlockLoRA, which leads to meaningful interactions among all merged concepts and significantly improves over existing instant merging methods.


\end{document}